\documentclass[runningheads]{llncs}
\usepackage{graphicx}
\usepackage{amsmath,amssymb} 
\usepackage{hyperref}
\usepackage{booktabs}
\usepackage{tabularx}
\usepackage{color}
\usepackage{xspace}
\usepackage{subcaption}
\usepackage[width=122mm,left=12mm,paperwidth=146mm,height=193mm,top=12mm,paperheight=217mm]{geometry}

 \makeatletter
 \DeclareRobustCommand\onedot{\futurelet\@let@token\@onedot}
 \def\@onedot{\ifx\@let@token.\else.\null\fi\xspace}

 \makeatother

\graphicspath{{./fig/}{./fig/plots/}}

\begin{document}

\pagestyle{headings}
\mainmatter
\definecolor{orange}{rgb}{1,0.5,0}
\definecolor{darkgreen}{rgb}{0,0.5,0}
\definecolor{red}{rgb}{1,0,0}

\newcommand{\modelname}{{\it Simple Feed Forward Network}\xspace}
\newcommand{\modelnameshort}{SFF\xspace}
\newcommand{\modulename}{{\it Multimodal Core}\xspace}
\newcommand{\modulenameshort}{MC\xspace}

\title{The Visual QA Devil in the Details: The Impact of Early Fusion and Batch Norm on CLEVR
\vspace{-0.5cm}}

\titlerunning{The Visual QA Devil is in the Details}

\author{Mateusz Malinowski\and
Carl Doersch 
}
%
\authorrunning{M. Malinowski, and C. Doersch}
%

\institute{\vspace{-0.3cm}DeepMind, London, United Kingdom \\
}

\maketitle
\vspace{-0.5cm}
\section{Introduction}
Visual QA
is a pivotal challenge for higher-level reasoning~\cite{malinowski14visualturing,geman2015visual,malinowski2017towards,agrawal2017don}, requiring understanding language, vision, and relationships between many objects in a scene.
Although 
datasets like CLEVR~\cite{johnson2017clevr} are designed to be unsolvable without such complex relational reasoning, some surprisingly simple feed-forward, ``holistic'' models have recently shown strong performance on this dataset~\cite{santoro2017simple,perez2017film}.  
These models lack any kind of explicit iterative, symbolic reasoning procedure, which are hypothesized to be necessary for counting objects, narrowing down the set of relevant objects based on several attributes, etc. 
The reason for this strong performance is poorly understood. 
Hence, our work analyzes such models, and finds that minor architectural elements 
are crucial to performance.
In particular, we find that \textit{early fusion} of language and vision 
provides large performance improvements. 
This contrasts with the late fusion approaches 
popular at the dawn of Visual QA~\cite{johnson2017clevr,malinowski2017ask,ren2015image,yang2015stacked}.
We propose a simple module we call \modulename (\modulenameshort), which we hypothesize performs the fundamental operations for multimodal tasks.  
We believe that understanding why these elements are so important to complex question answering will aid the design of better-performing algorithms for Visual QA while minimizing hand-engineering effort.

\vspace{-0.2cm}
\section{Models}
%
\begin{figure}[!b]
\begin{center}
\includegraphics[width=0.75\linewidth]{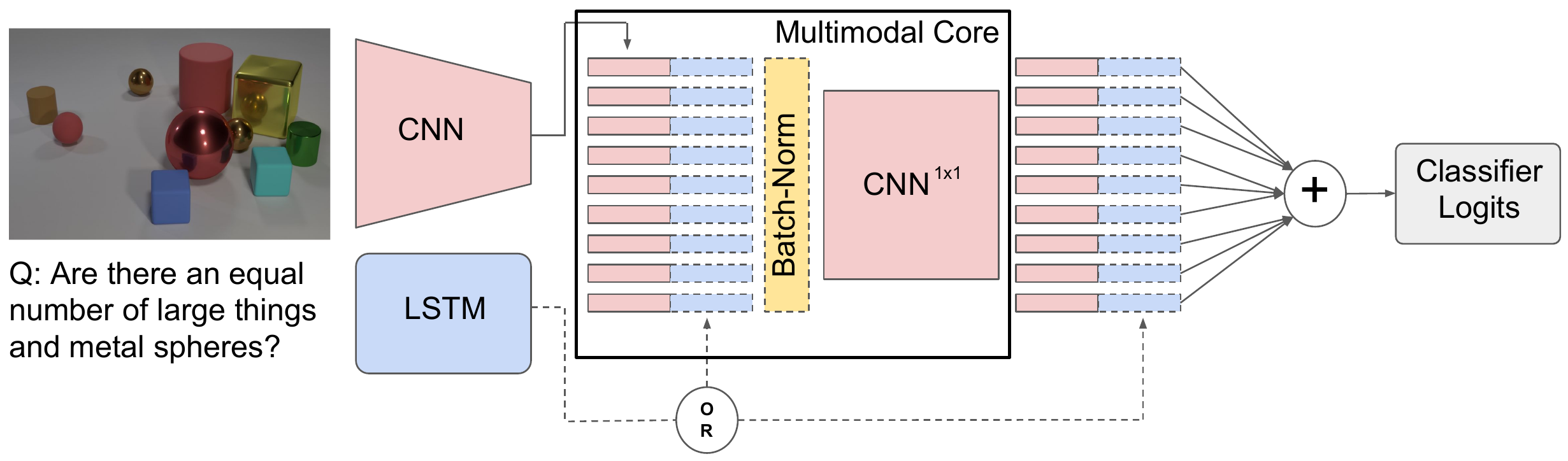}
\end{center}
\vspace{-0.7cm}
\caption{Our \modelname. The dotted lines denote exchangeable modules that their presence/absence we investigate wrt. the overall performance.}
\label{fig:clevr_net}
\vspace{-0.8cm}
\end{figure}
\noindent 
\autoref{fig:clevr_net} summarizes our architectures.
We use and image CNN with 4 layers of convolutions, ReLU, 128 kernels of size 3-by-3 and strides 2, and batch-norm.
The output tensor is $16\times16\times128$. 
We use an LSTM to process questions into vectors of size 128. 
The LSTM output is copied to all spatial locations in the CNN and concatenated with the image tensor.
As depicted in \autoref{fig:clevr_net}, we investigate two variants for fusing the vision and language. 
The first, which we term early+\modelnameshort, fuses the question vector with the image tensor, and then processes the fused representation by 4 layers of 1-by-1 convolutions (CNN$^{1x1}$), each with ReLU, 256 kernels, strides 1.
The second architecture, which we term late-\modelnameshort, processes the image tensor with the CNN$^{1x1}$, without language, only fusing the question vector afterward. 
We also experiment with an optional batch-norm layer just before CNN$^{1x1}$. 
We use sum-pooling to aggregate the resulting spatial tensor, with no  attention~\cite{yang2015stacked,malinowski2018han,xu2015ask}.
The remaining configuration is analogous to~\cite{santoro2017simple,malinowski2018han}.
Thus, we have 4 variants of our architecture: early+batch+\modelnameshort, late+batch+\modelnameshort, early+\modelnameshort, and late+\modelnameshort. 
Since empirically early+batch+\modelnameshort performs best, we name its core  \modulename (\modulenameshort), consisting of  multimodal fusion, followed by batch-norm, followed by CNN$^{1x1}$.

\vspace{-0.2cm}
\section{Experiments and Conclusions}
\autoref{table:results} summarizes our findings. 
The top (first 4 rows) shows overall, and per-question-type, performance of our variants of \modelnameshort. 
Early fusion is indeed critical for CLEVR; batch-norm shows small but non-trivial gains, especially on harder questions like counting or comparing numbers. 
We see a drastic improvement of using \modulenameshort relative to SAN~\cite{johnson2017clevr,yang2015stacked},
suggesting that SAN processes the multimodal features too late in the pipeline, and too ``shallowly''. 
Although late fusion may work on more biased Visual QA datasets~\cite{malinowski2017ask,ren2015image,malinowski14nips,antol2015vqa}, where exploiting language  biases plays a more prominent role, we believe the early fusion is more important on less biased datasets~\cite{agrawal2017don,johnson2017clevr}. 
To explore further, we implemented a late fusion variant of RN~\cite{santoro2017simple}. 
We train these models with $16\times16\times128$ image tensor, following the ``hard attenion'' pipeline~\cite{malinowski2018han}.
We call the reference model~\cite{malinowski2018han} early+batch+HAN+RN. The results (6th and 7th rows)  confirm that early fusion is crucial even in the relational models. 

So what is important for CLEVR performance?
CLEVR high-performing models like RN and FiLM differ from earlier models by performing early fusion.
In RN, multimodal features are processed by pairwise terms followed by 4 layers of CNN$^{1x1}$ ($g_\theta$ in~\cite{santoro2017simple}).
FiLM replaces concatenation with addition and multiplication that modulates batch-norm, and re-uses the module iteratively and residually~\cite{he2015deep}, which suggests that fusion doesn't need to be concatenation, but does need to happen early. 
Our work shows that simpler models also benefit enormously from early fusion, leading to highly performing, feed forward and ``holistic'' CLEVR models. 
We believe \modulenameshort captures important ideas shared by existing highly performing CLEVR architectures~\cite{santoro2017simple,perez2017film,dumoulin2018feature-wise} that can potentially be used in other multimodal problems.

\begin{table*}[tb]
\begin{center}
\scalebox{0.80}{
\begin{tabular}{l|c|ccccc}
\toprule
  & Overall\,\, & Count\,\, & Exist \,\, & Compare \, & Query      \, & Compare \\
  &             &           &            & Numbers \, & Attributes \, & Attributes \\
 \cmidrule(l){1-1}\cmidrule(r){2-7}
 $\circ$ early+batch+\modelnameshort  &\textbf{95.5} & \textbf{91.0} & \textbf{98.5} & \textbf{84.7} & \textbf{98.4} & \textbf{98.7} \\
 $\cdot$\, early+\modelnameshort & 94.4 & 88.6 & 97.7 & 82.9 & 98.0 & 97.6 \\
  $\circ$ late+batch+\modelnameshort & 58.0 & 51.9 & 71.4 & 72.2 & 54.4 & 55.8 \\
 $\cdot$\, late+\modelnameshort & 56.3  & 51.0 & 72.6 & 71.3 & 50.9 & 55.0 \\
 \midrule
 SAN~\cite{johnson2017clevr,yang2015stacked}  & 68.5 & 52.2 & 71.1 & 73.5 & 85.3 & 52.3 \\
 early+batch+HAN+RN~\cite{malinowski2018han} & 98.8 & 97.2 & 99.6 & 96.9 & 99.6  & 99.6  \\
 late+batch+HAN+RN & 57.2 & 50.3  & 70.7 & 73.1  & 53.9  & 54.5 \\
\bottomrule
\end{tabular}
}
\end{center}
\vspace{-0.15cm}
\caption{
CLEVR. $\circ$ and $\cdot$ denote corresponding (early vs late) experiments.
}
\vspace{-0.3cm}
\label{table:results}
\end{table*}

\clearpage

\bibliographystyle{splncs}
\bibliography{biblioLong,egbib}
\end{document}